\documentclass{article}

\usepackage{arxiv}
\usepackage{comment}
\usepackage[utf8]{inputenc} 
\usepackage[T1]{fontenc}    
\usepackage{hyperref}       
\usepackage{url}            
\usepackage{booktabs}       
\usepackage{amsfonts}       
\usepackage{nicefrac}       
\usepackage{caption}
\usepackage{microtype}      
\usepackage{lipsum}
\usepackage{graphicx}
\usepackage{listings}
\graphicspath{ {./images/} }
\usepackage{mathtools}
\usepackage{multirow}
\usepackage{multicol}

\usepackage{listings}
\usepackage{xcolor}
\definecolor{darkgreen}{rgb}{0, 0.5, 0}
\title{TorchSISSO: A PyTorch-Based Implementation of the Sure Independence Screening and Sparsifying Operator for Efficient and Interpretable Model Discovery}

\author{
 Madhav Muthyala \\
  Chemical and Biomolecular Engineering\\
  The Ohio State University\\
   Columbus, OH, USA \\
   \And
 Farshud Sorourifar \\
  Chemical and Biomolecular Engineering\\
  The Ohio State University\\
   Columbus, OH, USA \\
  \And
 Joel A. Paulson \\
  Chemical and Biomolecular Engineering\\
  The Ohio State University\\
   Columbus, OH, USA \\
  Correspondence: \texttt{paulson.82@osu.edu} \\
}
\hypersetup{
pdftitle={TorchSISSO: A PyTorch-Based Implementation of the Sure Independence Screening and Sparsifying Operator for Efficient and Interpretable Model Discovery},
pdfsubject={cs.LG},
pdfauthor={Madhav Muthyala, Farshud Sorourifar, Joel A. Paulson},
pdfkeywords={symbolic regression, interpretable machine learning, SISSO},
}

\DeclareMathOperator*{\argmin}{argmin}

\begin{document}
\maketitle

\begin{abstract}
Symbolic regression (SR) is a powerful machine learning approach that searches for both the structure and parameters of algebraic models, offering interpretable and compact representations of complex data. Unlike traditional regression methods, SR explores progressively complex feature spaces, which can uncover simple models that generalize well, even from small datasets. Among SR algorithms, the Sure Independence Screening and Sparsifying Operator (SISSO) has proven particularly effective in the natural sciences, helping to rediscover fundamental physical laws as well as discover new interpretable equations for materials property modeling. However, its widespread adoption has been limited by performance inefficiencies and the challenges posed by its FORTRAN-based implementation, especially in modern computing environments. In this work, we introduce TorchSISSO, a native Python implementation built in the PyTorch framework. TorchSISSO leverages GPU acceleration, easy integration, and extensibility, offering a significant speed-up and improved accuracy over the original. We demonstrate that TorchSISSO matches or exceeds the performance of the original SISSO across a range of tasks, while dramatically reducing computational time and improving accessibility for broader scientific applications.
\end{abstract}

\section{Introduction}
First principles models, derived from fundamental physical laws, have been instrumental in the development of scientific theories and technological systems. For example, the Navier-Stokes equation offers a comprehensive description of fluid flow, enabling predictions of complex behaviors in everything from blood flow~\cite{PESKIN1972252} to weather patterns~\cite{PHILLIPS196043}. Traditionally, this pursuit has relied on the extensive expertise of domain specialists, requiring trial and error to identify features and model structures that fit the observations. In recent years, the landscape of scientific inquiry has been transformed by the availability of machine learning frameworks, such as neural networks, support vector machines, and Gaussian processes, which offer a powerful alternative for deriving predictive models \cite{Karthikeyan2021}. These data-driven regression methods are often complex, do not typically generalize outside of the training set, and provide limited insights into the underlying physics. For instance, while these models may be trained to accurately predict the Reynolds number, they cannot capture the competitive nature between inertial and viscous forces in fluid flow. The only data-driven modeling framework that can provide insights comparable to first principles models, to the best of our knowledge, is symbolic regression (SR) \cite{Koza1994, wang2019symbolic, la2021contemporary}.

SR is an automated supervised learning technique that takes a user provided operator set and initial feature space to engineer expressions by combinatorically applying the operators to the base features set.
Early work in SR \cite{Koza1994} introduced the concept of using genetic programming (GP) to discover mathematical expressions and computer programs. The framework evolves a population of mathematical equations by applying genetic operations to the fittest individuals from the space of engineered expressions. 
Building on this work, Eureqa \cite{Eureqa}, developed a fitness function used to evaluate and evolve the population towards a ground-truth model. The GPLearn algorithm \cite{gplearn} is an open source implementation that improved on Eureqa by adding custom operators and the option to include constraints. 
The AI-Feynman and subsequent AI-Feynman 2.0 \cite{AIFeynman,AIFeynman2} build on this work by first exploiting simplifying properties of the data to to improve reliability and second returning a Pareto-optimal set of models to balance the model complexity with accuracy. Most recently, PySR \cite{PySR} has proposed several modifications to the genetic-based SR frameworks. This work proposed the use of a simulated annealing to actively tune the the fitness function used for identifying the fittest individuals from the population, a model simplifying stage between evolving candidates and optimizing the model parameters, and incorporates a novel complexity metric as a penalty in the fitness function.  

The approaches discussed thus far employ creative strategies to navigating the enormous spaces of possible models, due to high computation demand of exhaustive exploration. However, the these approaches are not guaranteed to find the correct model structure, as SR  has been proven to be an NP-hard problem \cite{virgolin2022symbolicregressionnphard}. While a truly exhaustive search would not be possible, several methods have investigated strategies to perform a targeted search over the sparse models. The Sparse Identification of Nonlinear Dynamics (SINDy) method \cite{SINDy} uses traditional sparse regression methods over an engineered feature space to balance model complexity with prediction accuracy, mainly for dynamic systems. An important challenge with SINDy in practice is the selection of the pre-defined feature set that plays big role in the achievable performance (e.g., the method will start to struggle if too many expanded features are considered). The Sure Independence Screening and Sparsifying Operators (SISSO) method \cite{SISSO} instead aims to tackle the problem of working with huge feature spaces (up to $\sim 10^9$ candidate features) by combining a fast feature screening method with exhaustive search over the subspace of features. SISSO relies on sure independence screening (SIS) \cite{SIS} to identify the most correlated features to the target using a simple dot product and a sparsity operator (typically $\ell_0$ regularization) to find the best simple model that fits the available training data. Recent work has also shown that SISSO can effectively be combined with other feature screening methods, such as mutual information pre-screening, to help deal with problems involving a large number of primary features/inputs before expansion \cite{iSISSO}. Furthermore, a Python wrapper package, \texttt{pysisso}, was recently developed to make the FORTRAN-SISSO implementation accessible to practitioners without knowledge of the FORTRAN language \cite{pySISSO}. However, the backend of \texttt{pysisso} still requires the a FORTRAN compiler, which does not fully address the difficulties with installation.

In this work, we present the \texttt{TorchSISSO} package, a user-friendly Python implementation of the SISSO framework designed to make the methodology accessible to a wider range of researchers and practitioners across diverse scientific fields. By eliminating the need for a FORTRAN compiler, \texttt{TorchSISSO} simplifies installation and usage, especially in modern computing environments. Furthermore, it allows users to easily modify the feature expansion process, which is hard-coded in the original FORTRAN implementation. This flexibility is a critical improvement, as we observed that the original SISSO does not always expand features as intended. Through simple examples, we demonstrate that \texttt{TorchSISSO} is capable of discovering the correct symbolic expressions in cases where the FORTRAN-based version cannot.

Additionally, the combinatorial expansion of the feature space may be slow or even infeasible, depending on the available memory. To address this issue, \texttt{TorchSISSO} uses parallel computing and optional GPU acceleration, providing significant computational speed up and scalability of the SISSO method.
The remainder of the manuscript is organized as follows: first, we provide a detailed description of the SISSO framework in Section \ref{sec:background}, and introduce the proposed toolbox in Section \ref{sec:torchsisso}. In Section \ref{sec:comparison}, we present performance comparison metrics for the proposed \texttt{TorchSISSO} to the \texttt{FORTRAN-SISSO} package. Lastly, we provide concluding remarks in Section \ref{sec:conclusion}.

\section{The SISSO Method}\label{sec:background}

The SR problem can be formulated as an empirical risk minimization over a function space $\cal F$. For given target variables $y^{(i)}\in\mathbb{R}$ and (base) feature variables $x^{(i)}\in\mathbb{R}^{d}$ for $i\in \{ 1,...,N \}$ data points, the SR problem can be defined as \cite{makke2024interpretable}
\begin{align} \label{eq:sr-problem}
    f^\star = \argmin_{f \in \mathcal{F}} \frac{1}{N}\sum_{i=1}^{N} L( f(x^{(i)}), y^{(i)} ).
\end{align}
The target (also known as outcome or response) represents the dependent variable that is assumed to be a deterministic transformation of a set of base features (also known as inputs) that are the independent variables in the problem. Here, $\mathcal{F}$ consists of all possible mappings $f : \mathbb{R}^{d} \to \mathbb{R}$ (needs to be specified by the user) and $f^\star$ is the optimal model that produces the lowest average loss $L(\cdot)$, across the training data.

The main difference between classical regression methods and SR is how $\cal F$ is defined. Classical regression defines the function space by assuming a structural form $\mathcal{F} = \{ f_\theta(x), ~ \forall \theta \in \Theta\}$, where $\theta$ is a collection of model parameters in some set $\Theta$. As long as the structures $f_\theta$ lead to differentiable loss functions in \eqref{eq:sr-problem}, one can then apply (stochastic) gradient descent methods to approximately solve \eqref{eq:sr-problem} (to at least a local optimum depending on the convexity of the loss). The SISSO method, on the other hand, aims to optimize over a set $\mathcal{F}$ that is formed by function composition over a primitive set. The primitive can contain variables, algebraic operators (such as addition, subtraction, multiplication), and transcendental functions (such as exponential, square root). The set $\mathcal{F}$ then contains all valid combinations of elements of the primitive applied recursively up until some level (see, e.g., \cite{virgolin2022symbolicregressionnphard} for details). A key challenge with this perspective is that the size of $\mathcal{F}$ grows exponentially fast with the size of the primitive set, and this space is finite (for fixed recursion depth), such that solving \eqref{eq:sr-problem} exactly requires exhaustive brute force search over all functions in $\mathcal{F}$. 

SISSO can be thought of as an effective heuristic to exactly search over a subset of useful functions in $\mathcal{F}$. We break down our description of SISSO, originally proposed in \cite{SISSO}, into three parts. First, we describe how feature expansion is recursively performed to build an engineered feature set that in general will be a subset of $\mathcal{F}$. Second, we summarize the sure independence screening (SIS) that identifies the very small subset of features that we want to more carefully analyze. Lastly, we present the sparsifying operator (SO) component that shows how the best functional form is selected from the subset of features identified in the previous step.

\subsection{Feature Space Expansion}

The choice of $\mathcal{F}$ is completely up to the user, however, in general it will contain potentially too many functions to even store in memory. Therefore, SISSO aims to recursively build a set of ``expanded features'' by applying a set of operators to all possible combinations of features. We highlight that different choices of operators and base features (inputs) will lead to different constructions of $\mathcal{F}$. The goal of SISSO is then to efficiently build and search over a large number of expanded features for modeling the target variable.
Let $\boldsymbol{\phi}_0 = x$ be the initial features and let $\mathcal{O}$ denote the operator set that consists of some number of unary $o[x_i]$ and binary $o[x_i, x_j]$ operators. Then, we define the expanded features at level $l \geq 1$ recursively as follows
\begin{align} \label{eq:feature-exp}
    \boldsymbol{\phi}_l = \{ \mathcal{O}[ z_i, z_j ], ~~ \forall z_i, z_j \in \phi_{l-1} \} ~~~ \text{with} ~~~ \boldsymbol{\phi}_0 = x. 
\end{align}
As an example, consider the $d=2$ and a very simple operator set of $\mathcal{O} = \{ I(z_i), z_i + z_j, z_i \times z_j \}$. Then, we can construct the features up until level 2 as follows:
\begin{align*}
    \boldsymbol{\phi}_0 &= \{ x_1,~x_2 \}, \\
    \boldsymbol{\phi}_1 &= \{ x_1,~x_2,~x_1 + x_2,~x_1 \times x_2 \}, \\
    \boldsymbol{\phi}_2 &= \{ x_1, ~ x_2, ~ x_1 + x_2, ~ x_1 \times x_2, ~ 2x_1 + x_2, ~ x_1 + x_1 \times x_2, ~ x_1 + 2x_2, ~ x_2 + x_1 \times x_2, \\\notag
    & ~~~~~~~~~ x_1(x_1 + x_2), ~ x_1^2 \times x_2, ~ x_2(x_1 + x_2), ~ x_1 \times x_2^2, ~ (x_1 + x_2 + x_1 \times x_2), ~ (x_1 + x_2)(x_1 \times x_2) \}. 
\end{align*}
For $m_u$ unary operators, $m_{b,s}$ symmetric binary operators, and $m_{b,ns}$ non-symmetric binary operators, we can compute an upper bound on the number of features at any level $l \geq 1$:
\begin{align}
    d_l \leq m_u d_{l-1} + \left( \frac{m_{b,s}}{2} + m_{b,ns} \right) d_{l-1}(d_{l-1}-1), ~~ d_0 = d.
\end{align}
Note that this is an upper bound since it is possible that some of the combinations are not unique. In the example above, we get $d_1 \leq 2 + \left( \frac{2}{2} \right) (2) (1) = 4$, which is exact for the first level since all combinations are unique. For the second level, we get $d_2 \leq 4 + \left( \frac{2}{2} \right) (4) (3) = 16$. This bound is larger than the 14 unique combinations shown above because we can exclude, e.g., $x_1 + x_2$ and $x_1 \times x_2$ that would be regenerated when expanding from level 1 to 2.

The quadratic term quickly dominates as $l$ increases such that we can write out a rough scaling law as $d_l \sim m_b' d_{l-1}^2$ where $m_b' = (m_{b,s}/2 + m_{b,ns})$ for $l \geq 1$. Rewriting this in terms of the number of primary/starting input features, we find that the size of $\phi_l$ should be roughly
\begin{align}
    d_l \sim (m_b')^{2^l - 1} d^{2^l},
\end{align}
which grows exponentially with the number of levels $l$ (and the number of binary operators in the operator set). In practice, we can limit this growth by performing dimensional analysis during the expansion process, which restricts certain operators from being applied (e.g., addition and subtraction can only be applied if the features share the same units). However, this does place a strong limit on the maximum expansion level in SISSO -- typically needs to be below 4, except in special cases. Also, note that our implementation, described in Section \ref{sec:torchsisso}, enables the user significant flexibility in their choice of operator set $\mathcal{O}$, which plays a major role on the growth in the feature space. 
In practice, one may treat $l$ as a tunable hyperparameter, as the best choice will depend on the problem. Given the steep increase in cost as $l$ increases, we recommend a greedy approach: start with small $l$ (likely $l=1$) and incrementally increase it until model accuracy meets the requirements of the application. If one is unable to find a sufficiently good model with $l \leq 3$, it is recommended to revisit the choice of operators included in $\mathcal{O}$ before attempting even higher expansion levels.

\subsection{Sure Independence Screening}

Although higher expansion levels create a richer feature space for mapping the target, they also increase the complexity of the learning task. Specifically, finding an optimal sparse linear combination of these features becomes crucial to avoid overfitting, particularly in high-dimensional spaces. Sparsity is often achieved by applying regularization techniques in the regression process. Common strategies include $\ell_1$ regularization (LASSO) or a combination of $\ell_1$ and $\ell_2$ regularization (elastic net), which penalize non-zero coefficients to enforce sparsity in the model. However, selecting the appropriate hyperparameters (penalty weights) can be both challenging and time-consuming. This issue is particularly pronounced in limited data settings, where extensive validation to tune these hyperparameters is often infeasible, leading to potential model instability. The SISSO method tackles this problem by first applying sure independence screening (SIS) \cite{SIS} to quickly and efficiently select a much smaller set of features for use in the modeling training/selection step. 

SIS is a simple, non-parametric statistical method designed for variable selection in high-dimensional feature spaces. Variables are ranked based on the correlation magnitude metric between each feature and the target. Let $\boldsymbol{y} \in \mathbb{R}^N$ be the vector of training target values and $\boldsymbol{\Phi} \in \mathbb{R}^{N \times D}$ be matrix of feature values that corresponds to all $D$ features evaluated at the $N$ training input values. Note that we describe the SIS procedure for an arbitrary feature matrix that could be derived from any expansion level. Assuming the columns of $\boldsymbol{\Phi}$ have been standardized to have zero mean and unit variance, we can compute the following weights that measure the correlation between each feature and the target:
\begin{align} \label{eq:sis-weight}
    \boldsymbol{w} = (w_1, \ldots, w_D) = \boldsymbol{\Phi}^\top \boldsymbol{y}.
\end{align}
SIS then identifies the indices (the particular features) with the top $k$ magnitude weight:
\begin{align} \label{eq:sis-method}
    \mathcal{S} = \{ i \in \{ 1,\ldots, D \} : | w_i | ~\text{is among the first $k$ largest} \}.
\end{align}
We denote this process with the shorthand: $\mathcal{S} = \text{SIS}( \boldsymbol{y}, \boldsymbol{\Phi} )$. Note that the choice of $k$ is up to the user; larger values will make the subsequent step more computationally demanding. We implement a default value of $k=20$ based on the recommendation from \cite{SISSO}. An alternative strategy is to only keep features whose correlation $w_i$ exceed a threshold value, which is also implemented in our \texttt{TorchSISSO} package. 

\subsection{Sparsifying Operator}

Let $\boldsymbol{\phi}(x)$ denote the set of nonlinearly expanded features at any expansion level (we suppress the subscript $l$ for notational simplicity). We are aiming to find a model that is a linear combination of these features, i.e., $\boldsymbol{\phi}(x)^\top \boldsymbol{c}$ where $\boldsymbol{c} \in \mathbb{R}^{D}$ is a coefficient vector that we want to fit to data. Note that we assume the constant feature is included in $\boldsymbol{\phi}(x)$ to serve as a bias term in the model. Although we could fit $\boldsymbol{c}_l$ using standard linear regression, this problem will be underdetermined when $D > N$, which is typically the case. We also do not expect the vast majority of the features to be important when predicting $y$. SISSO thus combines SIS with a sparsifying operator (SO) to overcome this challenge. 

Let $\boldsymbol{\Phi}_{\mathcal{S}} \in \mathbb{R}^{N \times k}$ denote the submatrix of feature matrix $\boldsymbol{\Phi}$ that extracts columns with indices $\mathcal{S}$. Since $k \ll D$, it is now typically possible to use standard linear regression to fit the coefficients of the $k$ remaining features. However, it is still not clear how many non-zero coefficients to retain in the model. We could address this problem using more traditional regularization methods mentioned previously, but this introduces some additional tuning parameters that are hard to select in practice. SISSO takes an alternative approach to address this issue by sequentially building models from a single term (one feature/descriptor) up until a maximum number of $T$ terms. Every time that a new term is considered, the residual error from the previous model is used to guide the chioce of the feature subset. Let $\boldsymbol{r}_t \in \mathbb{R}^N$ denote the residual error for a model with $t$ terms selected from a subset $\mathcal{S}_t$. It turns out that we can compute $\boldsymbol{r}_t$ in closed form as follows
\begin{align} \label{eq:residual}
    \boldsymbol{r}_t = \boldsymbol{y} - \boldsymbol{\Phi}_{\mathcal{S}_t} \boldsymbol{E}_t \boldsymbol{c}_t ~~~ \text{where} ~~~ \boldsymbol{c}_t = ( \boldsymbol{E}_t^\top \boldsymbol{\Phi}^\top_{\mathcal{S}_t} \boldsymbol{\Phi}_{\mathcal{S}_t} \boldsymbol{E}_t )^\top \boldsymbol{E}_t^\top \boldsymbol{\Phi}^\top_{\mathcal{S}_t} \boldsymbol{y},
\end{align}
where $\boldsymbol{E}_t \in \mathbb{R}^{K \times t}$ is a binary matrix that selects $t$ feature columns out of the available ones in $\boldsymbol{\Phi}_{\mathcal{S}_t} \in \mathbb{R}^{N \times K}$, $K$ is the number of features in $\mathcal{S}_t$, and $\boldsymbol{c}_t \in \mathbb{R}^t$ is the coefficient vector corresponding to the least squares solution from fitting $\boldsymbol{\Phi}_{\mathcal{S}_t} \boldsymbol{E}_t$ to $\boldsymbol{y}$. Furthermore, let $\boldsymbol{r}_t^\star$ denote the residual error for the best model tested with $t$ terms from the subspace $\mathcal{S}_t$. SISSO recursively adds more features to the subspace as follows
\begin{align} \label{eq:subspace-expansion}
    \mathcal{S}_{t+1} = S_t \cup \text{SIS}( \boldsymbol{r}_t^\star, \boldsymbol{\Phi} ) ~~~ \text{with} ~~~ \mathcal{S}_1 = \text{SIS}( \boldsymbol{y}, \boldsymbol{\Phi} ).
\end{align}
In words, this procedure looks at the best $t$-term residual and then adds the next $k$ best features with the highest SIS scores with respect to the residual. We can actually compute $\boldsymbol{r}_t^\star$ using exact $\ell_0$ regression (or exhaustive search) over all possible $t$ term models in $\mathcal{S}_t$, which corresponds to the minimum $\| \boldsymbol{r}_t \|^2$ over $\binom{tk}{t}$ models. The SISSO method keeps executing \eqref{eq:subspace-expansion} until the best model found for a particular $t$ achieves low enough error or until the maximum number of terms $T$ is reached. Since the number of trained models grows quickly with $T$, we typically set it to be $T=3$, meaning we at most consider 3 term models (though again this choice can be easily modified by users in our implementation). This means SISSO will attempt at most $\sum_{t=1}^T {\binom{tk}{t}}$ least square regression steps. The best trained model (i.e., the model with the lowest residual norm) is returned as the final model. 

\begin{figure}[th]
    \centering
    \includegraphics[width = \linewidth]{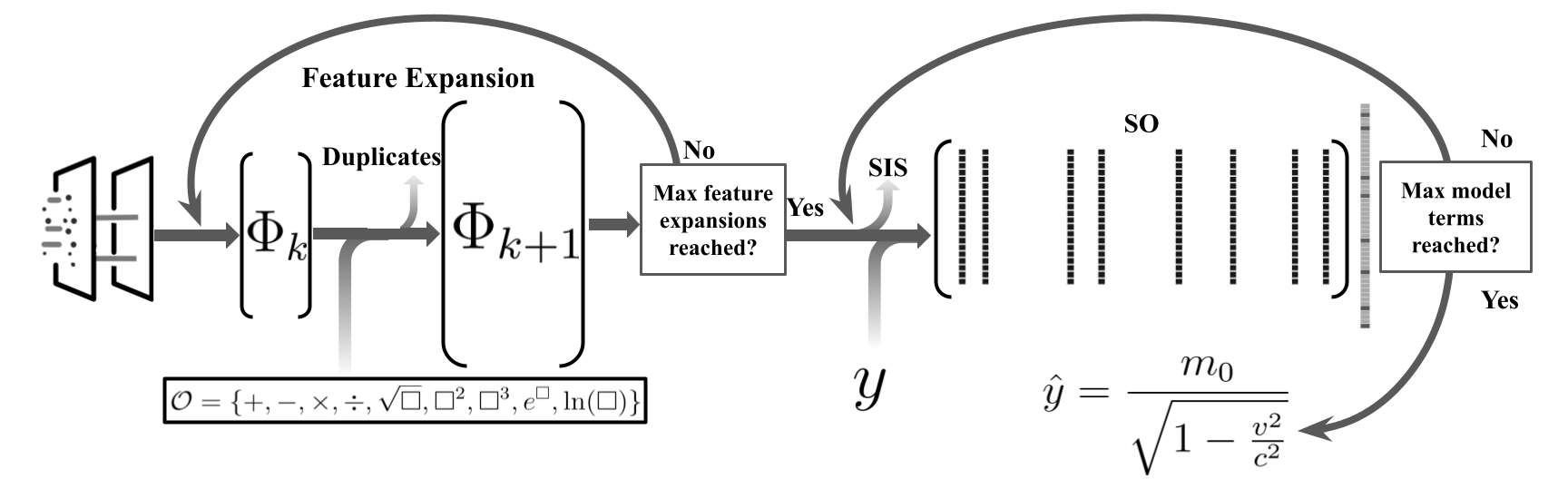}
    \caption{Illustration of the major steps in the SISSO method from \protect{\cite{SISSO}}.}
    \label{fig:sisso-illustration}
\end{figure}

A simple illustration of the complete SISSO method is shown in Figure \ref{fig:sisso-illustration}. 

\subsection{Impact of Noise and Data Distribution}

The SISSO framework assumes that the target variable is a (sparse) linear combination of the expanded features, with observations potentially corrupted by additive random noise:
\begin{align} \label{eq:target-noise}
    y^{(i)} = \boldsymbol{\phi}(x^{(i)})^\top \boldsymbol{c} + \varepsilon^{(i)}, \quad i = 1, \ldots, N,
\end{align}
where each $\varepsilon^{(i)}$ is an unobserved error term assumed to follow a given probability distribution. The assumed distribution of $\varepsilon^{(i)}$ influences the choice of loss function in \eqref{eq:sr-problem}. Currently, our implementation focuses on the commonly used “least squares” formulation, derived from maximum likelihood estimation (MLE) under the assumption that the noise $\{ \varepsilon^{(i)} \}_{i=1}^N \sim \mathcal{N}(0, \sigma^2)$ follows an i.i.d. zero-mean Gaussian distribution.

In cases where the noise vector $\boldsymbol{\varepsilon} = (\varepsilon^{(1)}, \ldots, \varepsilon^{(N)})$ instead follows a more general multivariate Gaussian distribution, $\boldsymbol{\varepsilon} \sim \mathcal{N}(\mathbf{0}, \boldsymbol{\Sigma})$, a whitening transformation can be applied to map the problem to a space where the noise terms are i.i.d., provided $\boldsymbol{\Sigma}$ is known. Specifically, we compute a weight matrix $\boldsymbol{W} = \boldsymbol{\Sigma}^{-1/2}$ and define transformed targets and features, $\tilde{\boldsymbol{y}} = \boldsymbol{W} \boldsymbol{y}$ and $\tilde{\boldsymbol{\Phi}} = \boldsymbol{W} \boldsymbol{\Phi}$, satisfying the i.i.d. noise assumption. After SISSO identifies a model in the transformed space, this result can be converted back to the original scale as needed for prediction and interpretation.

The level of noise in the observations can significantly affect the ability to accurately learn the governing equations. In practice, given finite data, the best-fit models will depend on interrelated factors such as the number of data points, the noise level, the structure of the ground-truth equations, and the range of the observed data. Since these dependencies are often difficult to understand \textit{a priori}, standard practice in machine learning emphasizes extensive validation and testing. Validation on held-out test data is one straightforward approach, but for low-data settings, $k$-fold cross-validation is generally preferred as it leverages the full dataset while providing an estimate of performance variation across subsets of the data.
Regardless of the validation method, users should remain cautious of the “best” model identified by SISSO or any other algorithm until it has been thoroughly evaluated in tasks that reflect its intended application. Robustness to noise can also be enhanced by ensembling techniques, such as bootstrap aggregation (bagging) \cite{sagi2018ensemble}, which has been shown to improve model resilience, as demonstrated with the SINDy algorithm in \cite{fasel2022ensemble}. Similar approaches may be applied to SISSO, a direction we intend to pursue in future work.

Additionally, the distribution of the input variables (base features $x$) plays a critical role in the model learned. For instance, if all measurements are concentrated within a narrow range in the input space, the data will contain limited information about the overall target variable distribution, and consequently, learning a broadly accurate model becomes nearly impossible. Section \ref{subsec:extrapolation} provides an illustrative example showing how a model can achieve high accuracy within a localized region of the input space, highlighting the importance of validation, particularly when extrapolating beyond the observed data range.

\section{The TorchSISSO Package}\label{sec:torchsisso}

\subsection{Feature Pre-screening for High-dimensional Problems} \label{subsec:mi-screening}

The original version of SISSO, outlined in Section \ref{sec:background}, does not scale to high-dimensional primary features $x \in \mathbb{R}^d$, i.e., when $d$ is very large. Since this case commonly arises in practical applications (e.g., molecular property modeling), we incorporate a strategy for dealing with large $d$ in \texttt{TorchSISSO}. Specifically, we implement an optimal mutual information (MI) screening procedure that has been previously explored in \cite{iSISSO, battiti1994using}. MI between a component of the primary feature vector $x_i$ and the target $y$ is defined as
\begin{align} \label{eq:mi-exact}
    \text{MI}(y ; x_i) = \int p(x_i, y) \log\left( \frac{p(x_i, y)}{p(x_i) p(y)} \right)\text{d}x_i \text{d}y, 
\end{align}
where $p(x_i, y)$ is the joint probability density function between $x_i$ and $y$, $p(x_i)$ is the marginal probability density function of $x_i$, and $p(y)$ is the marginal probability density function of $y$. MI is a strictly non-negative measure of the relationship between $x_i$ and $y$ and is only zero if $x_i$ and $y$ are statistically independent. In practice, we approximate the integral in \eqref{eq:mi-exact} with kernel density estimation. MI is used to down-sample the feature space, effectively assuming that high MI implies higher likelihood that a feature contributes to the target prediction. Based on the choice by the user, we either keep the top ranked MI features up until a maximum number of terms or keep only the features whose MI fall into a specified quantile range. 

\subsection{PyTorch Implementation}

To ensure a flexible and easy to use/install package, we decided to implement the SISSO algorithm in \texttt{PyTorch} \cite{paszke2019pytorch}, which is an open-source machine learning library. A key feature of \texttt{PyTorch} is its Tensor computing framework that allows efficient implementation of multivariate tensor objects with strong acceleration using, e.g., graphics processing units (GPUs). This makes it straightforward to efficiently carry out the most expensive operations in SISSO. Looking back at Section \ref{sec:background}, we see that SISSO mainly involves performing recursive feature expansion \eqref{eq:feature-exp}, running SIS via the matrix-vector multiplication in \eqref{eq:sis-weight}, and fitting many models with a small number of terms to find the residuals in \eqref{eq:subspace-expansion}. All of these steps can be straightforwardly executed using native operations in \texttt{PyTorch}. For feature expansion, \texttt{PyTorch} is highly optimized to perform efficient element-wise operations on tensors, which can be executed in parallel, leveraging the available power of the CPU or GPU for fast computation. In addition to supporting a wide variety of element-wise operations, \texttt{PyTorch} also enables broadcasting the result to tensors of different shapes. 
The \texttt{torch.matmul} function for matrix multiplication is generally very efficient, especially for large matrices. This makes it straightforward to execute the SIS procedure, even as the feature matrix $\boldsymbol{\Phi}$ gets very large. Lastly, the \texttt{torch.linalg.lstsq} function can be used to efficiently compute the residual $\boldsymbol{r}_t$ for a $t$-term model. Unlike many existing linear least square methods, \texttt{torch.linalg.lstsq} can simultaneously solve a ``batch'' of problems. This means we can simultaneously solve \eqref{eq:residual} for all $\binom{tk}{t}$ possible models (the different possible binary matrices $\boldsymbol{E}_t$), as opposed to sequentially solving each problem within a standard \texttt{for} loop. Note that we do not explicitly construct $\boldsymbol{E}_t$ and multiply it by the feature matrix, as this would be inefficient. Instead, we broadcast all possible $t$-term combinations of the features into a $B \times t \times N$ tensor where $B = {\binom{tk}{t}}$ is the batch size (number of combinations of the $tk$ features split into $t$ terms), $t$ is the number of terms considered, and $N$ is the number of datapoints.

\subsection{Installation and Usage}

The \texttt{TorchSISSO} package can be installed using the PIP package manager as follows
\begin{quote}
\begin{lstlisting}
pip install TorchSisso
\end{lstlisting}
\end{quote}
endgroup
The complete package is available on Github, which includes a Google Colab notebook that implements a series of simple examples using \texttt{TorchSISSO} that can be run interactively in the cloud\footnote{The Github code to the \texttt{TorchSISSO} package can be found at this link \href{https://github.com/PaulsonLab/TorchSISSO}{https://github.com/PaulsonLab/TorchSISSO}. Fully worked out examples using \texttt{TorchSISSO} can be found at this link \href{https://colab.research.google.com/drive/1ObQJJXTpz5l04pphSH1nHT-Rsd2zBszC?usp=sharing}{https://colab.research.google.com/drive/1ObQJJXTpz5l04pphSH1nHT-Rsd2zBszC?usp=sharing}.}. All of the core operations of \texttt{TorchSISSO} can be accessed using the \texttt{SissoModel} class that can be imported as follows
\begin{quote}
\begin{lstlisting}
from TorchSisso import SissoModel
\end{lstlisting}
\end{quote}
To construct an instance of this class, one needs to set a number of inputs including a Pandas dataframe consisting of the training data \texttt{df}, the set of operators to include in the feature expansion step \texttt{operators}, the number of expansion levels \texttt{n\_expansion}, the number of terms in the final model \texttt{n\_term}, and the number of features to keep for every term in the model \texttt{k}. The first column of \texttt{df} should contain the target variable at all the training points $\boldsymbol{y}$ and the reminaing columns should contain the primary feature matrix $\boldsymbol{\Phi}_0 = \boldsymbol{X}$ that is expanded internally to form $\boldsymbol{\Phi}_l$ where $l$ is equal to \texttt{n\_expansion}. The operators should be passed in the form of a Python list, with each element being a string (for standard operators) or a function that can operate on \texttt{torch.Tensor} objects. We can then call the \texttt{.fit()} method to train the model, which returns the root mean squared error (RMSE) of the best-found model, a string version of the equation (that can easily be converted to symbolic form or a LaTeX expression), and the corresponding $R^2$ (coefficient of determination) value. Therefore, one can effectively train a model using SISSO with just a few lines of code:
\begin{quote}
\begin{lstlisting}
# import necessary packages
import numpy as np
import pandas as pd
from TorchSisso import SissoModel
# create dataframe with targets "y" and primary features "X"
data = pd.DataFrame(np.column_stack((y, X)))
# define unary and binary operators of interest
operators = ["+", "-", "*", "/", "exp", "ln", "pow(2)", "sin"]
# create SISSO model object with relevant user-defined inputs
sm = SissoModel(data, operators, n_expansion=4, n_term=1, k=5)
# run SISSO training algorithm to get interpretable model with highest accuracy
rmse, equation, r2 = sm.fit()
\end{lstlisting}
\end{quote}

There are two additional optional arguments that can be provided to \texttt{SissoModel} to help mitigate the growth of the feature space with number of expansion levels. The first is an \texttt{initial\_screening} argument that implements the MI screening approach described in Section \ref{subsec:mi-screening}. The data is passed as a list of the form \texttt{[method, quantile]} where \texttt{method="mi"} indicates the use of MI screening and \text{quantile} should be a floating point number between 0 and 1 that specifies only features with MI inside of this quantile range should be kept for expansion. We also implement a simple linear correlation pre-screening method, which can be selected by setting \texttt{method="spearman"}, though we typically find that MI performs better in practice. The second optional argument is \texttt{dimensionality} that should be a list of strings that represent the units of a given feature. For example, in the case that we have 5 features where features 1 to 4 have unique units while feature 5 shares the same units as feature 3, we would set this argument as \texttt{dimensionality = ["u1", "u2", "u3", "u4", "u3"]}. This ensures that non-physical features are not generated during the expansion process, reducing both memory usage and computational cost.

\section{Numerical Examples} \label{sec:comparison}

In this section, we compare the performance of \texttt{TorchSISSO} with the original SISSO implementation, referred to as \texttt{FORTRAN-SISSO}, and its derivatives across various test cases, including synthetic equations, challenging scientific benchmarks, and a real-world application in molecular property prediction. All results are based on a single realization of training data generated from the ground-truth equations, potentially corrupted by random observation/measurement noise. However, we found the results to be largely insensitive to the specific data realization. The experiments were run on a computing cluster with two nodes, each equipped with an Intel Xeon Gold 6444Y processor (16 cores) and 512 GB of DDR4 RAM.
All the experiments, except those in Section \ref{subsec:gpu-comparison}, are run on a CPU to provide a fair comparison with \texttt{FORTRAN-SISSO}, which currently does not provide GPU support.

\subsection{Synthetic Equations}
\label{subsec:synthetic}

We initially compare \texttt{TorchSISSO} to \texttt{FORTRAN-SISSO} on 10 synthetic expressions inspired from benchmarks commonly used in the symbolic regression (SR) literature \cite{makke2024interpretable}. The expressions are summarized in Table \ref{tab:synthetic}. For each expression, we generate 10 training datapoints by randomly sampling $x$ in $[1,5]^d$ where $d$ matches the number of variables appearing in the expression; all observations are corrupted with Gaussian noise with zero mean and standard deviation equal to 0.05. The computational time and the root mean squared error (RMSE) on the training set for the best-found models with \texttt{TorchSISSO} and \texttt{FORTRAN-SISSO} are shown in Table \ref{tab:synthetic}. 
We see that for several of the expressions (1, 2, 4, 6, 7, 8, 10), \texttt{TorchSISSO} obtains exactly the same RMSE as \texttt{FORTRAN-SISSO} but does so in less time. In the other three cases (3, 5, 9), \texttt{TorchSISSO} achieves low RMSE (indicating it has learned something very close to the ground-truth expression) while \texttt{FORTRAN-SISSO} learns a model with high RMSE (meaning it has failed to learn the ground truth). Case 5 is particularly interesting, as \texttt{TorchSISSO} finds a model with two orders of magnitude lower RMSE in nearly half the time (substantially improves in both metrics). It is not immediately obvious why \texttt{FORTRAN-SISSO} fails to learn the true structure for cases 3, 5, and 9; however, we believe this is due to some implementation differences in the feature expansion step. Regardless of the reason, \texttt{TorchSISSO} is clearly capable of achieving better performance in less time than the original \texttt{FORTRAN-SISSO}. 

\begin{table}[tb]
        \centering
        \caption{Ground-truth models for the synthetic equations and corresponding training time and RMSE for \texttt{TorchSISSO} and \texttt{FORTRAN-SISSO} on each equation. The \textbf{bold font} denotes a better score and the $^\star$ denotes a tied score.}
        \label{tab:time_and_accuracy_num}
        \begin{tabular}{|lll || ll || ll|}\hline
        & & &\multicolumn{2}{l||}{\hspace{.25in} \texttt{TorchSISSO} } &  \multicolumn{2}{l|}{\hspace{.15in} \texttt{FORTRAN-SISSO} } \\ \hline &&&&&& \\
        \# &\underline{\ Expression \ }& & \underline{\ Time (sec)\ }& \underline{\ RMSE\ } &  \underline{\ Time (sec)\ } & \underline{\ RMSE\ } \\  &&&&&& \\
        1&\multirow{2}{*}{$10\frac{x_1}{x_2(x_3+x_4)}$} & &	\textbf{0.04}&	0.0391$^\star$& 	0.11&	0.0391$^\star$\\ &&&&&& \\
        2&\multirow{2}{*}{$2\sin(x_2)+ 3\sqrt{x_1}$}& &	\textbf{0.01}&	0.0434$^\star$& 	0.32&	0.0434$^\star$\\ &&&&&& \\
        3&\multirow{2}{*}{$3\frac{\exp(x_1)}{x_2+\exp(x_3)}$}& &	0.26&	\textbf{0.0342}& 	\textbf{0.20}&	1.4359\\ &&&&&& \\
        4&\multirow{2}{*}{$3x_3+x_2^2+x_1^3$}& &	\textbf{0.27}&	0.0348$^\star$&  	0.57&	0.0348$^\star$\\ &&&&&& \\
        5&\multirow{2}{*}{$\frac{x_2+\exp(x_2)}{x_1^2-x_2^2}$}& &	\textbf{0.12}&	\textbf{0.0557}& 	0.22&	1.0786\\ &&&&&& \\
        6&\multirow{2}{*}{$\sqrt{x_1^2+x_2^2}$}& &	\textbf{0.02}&	0.0646$^\star$& 	0.29&	0.0646$^\star$\\ &&&&&& \\
        7&\multirow{2}{*}{$\sin(x_1x_3)+ 1.5\exp(-x_1x_2)$} & &	\textbf{0.00}&	0.0452$^\star$& 	0.39&	0.0452$^\star$\\ &&&&&& \\
        8&\multirow{2}{*}{$5(x_1x_3^3) + x_1^3 + 3(x_1x_2^2)$}& &	\textbf{0.01}&	0.0353$^\star$& 	0.27&	0.0353$^\star$\\ &&&&&& \\
        9&\multirow{2}{*}{$x_1x_2x_3\left(\ln(x_4)-\ln(x_5)\right) $}& &	66.96&	\textbf{1.61E-15}& 	\textbf{0.27}&	2.218\\ &&&&&& \\
        10&\multirow{2}{*}{$\exp(-\frac{x_1}{x_3x_2})$}& &	\textbf{0.04}&	1.17E-16$^\star$& 	0.12&	1.17E-16$^\star$\\ &&&&&& \\\hline
        \end{tabular}
\label{tab:synthetic}
\end{table}

\subsection{Scientific Benchmarks}
\label{subsec:scientific}

Next, we consider four equations from the SRSD-Feynman dataset \cite{matsubara2022rethinking}, which is a modified version of the data proposed in \cite{AIFeynman} to have more realistic sampling ranges for the primary features and constants. Each of these equations can be found in Richard Feynman's famous ``Lectures on Physics,'' and are becoming increasingly common as benchmarks for SR methods (because it mimics a realistic scientific task of discovering fundamental physical laws). The selected equations shown in Table \ref{tab:equations_phys} span a variety of physical phenomena including (i) the relationship between distance and two points in space, (ii) particle displacement in an electromagnetic field, (iii) relativistic mass as a function of velocity and the speed of light, and (iv) the oscillation amplitude of a charged particle in an electromagnetic field. We generate 50 training datapoints without noise using the distributions reported in \cite{matsubara2022rethinking}. The computational time and RMSE for both \texttt{TorchSISSO} and \texttt{FORTRAN-SISSO} are also shown in Table \ref{tab:equations_phys}. Note that we use dimensional analysis in both cases to limit the growth in the expanded feature set. We see that \texttt{TorchSISSO} achieves the best accuracy and, in fact, discovers the exact ground truth equation in all cases. \texttt{FORTRAN-SISSO}, on the other hand, is unable to derive the exact equation in any of the considered cases. It is worth noting that, for the final case (oscillation amplitude), \texttt{TorchSISSO} does take around 42 seconds as it requires going to a third expansion level. Although this is considerably longer than the other cases, this is still substantially less time than that required by most existing SR methods (that can take several hours to find expressions of similar complexity). 

\begin{table}[tb]
\centering
\caption{Ground-truth models for the scientific benchmarks and corresponding training time and RMSE for \texttt{TorchSISSO} and \texttt{FORTRAN-SISSO} on each equation. The \textbf{bold font} denotes a better score.}
\label{tab:equations_phys}
\begin{tabular}{|lll || ll || ll|}\hline
        & & &\multicolumn{2}{l||}{\hspace{.25in} \texttt{TorchSISSO} } &  \multicolumn{2}{l|}{\hspace{.15in} \texttt{FORTRAN-SISSO} } \\ \hline &&&&&& \\
         \underline{\ Name \ }& \underline{\ Physics-based Equation \ }& & \underline{\ Time (sec)\ }& \underline{\ RMSE\ } &  \underline{\ Time (sec)\ } & \underline{\ RMSE\ } \\  &&&&&& \\
        Distance&\multirow{2}{*}{$d^2 = (x_0-x_1)^2+(x_2-x_3)^2$}          & &	   {0.40}  & \textbf{1.35E-15}  &  \textbf{0.11} & 0.0363 \\ &&&&&& \\
        Particle Displacement &\multirow{2}{*}{$F=q(E + Bv\sin(\theta))$ }      & &	   \textbf{0.21}  & \textbf{ 2.1E-15}  &  {0.24} & 0.0449 \\ &&&&&& \\
        Relativistic Mass&\multirow{2}{*}{$m^2 = \frac{m_0^2 }{1 - \frac{v_1^2}{c^2}}$}    & &	   {1.44}  & \textbf{7.64E-6}   &  \textbf{0.13} & 1.185  \\ &&&&&& \\
        Oscillation Amplitude &\multirow{2}{*}{$x = \frac{qe} { m(\omega_1^2-\omega_2^2)} $}& &   42.25   & \textbf{6.31E-23}  &  \textbf{0.17}& 0.0402  \\ &&&&&& \\\hline
\end{tabular}
\end{table}

\subsection{Interpretable Models for Molecular Property Prediction}

As a final case study, we focus on constructing simple, interpretable models for predicting molecular properties -- an essential challenge in fields such as pharmaceuticals, materials science, and environmental science. Here, we look at modeling the specific energy of organic compounds, which is a property that is known to be strongly correlated to energy density when the material is used as an electrode in batteries \cite{tuttle2023predicting}. Specific energy can be computed using the following equation
\begin{equation} \label{eq:specific-energy}
\text{Specific Energy} = \frac{(E - E_{\text{anode}}) n F}{3600 M_W},
\end{equation}
where $E$ is the redox potential, $E_{\text{anode}}$ is the redox potential of the anode (in this case a Zinc anode), $n$ is the number of moles of electrons transferred, $F$ is Faraday's constant, and $M_W$ is the molecular weight of the molecule. 
All quantities in \eqref{eq:specific-energy} are known except for $E$, which can be approximated using density functional theory (DFT). The challenge, however, is that DFT is computationally expensive, making it impractical to scale \eqref{eq:specific-energy} to millions of candidate molecules. Larger candidate sets are essential when the goal is to discover multiple high-performance molecules. To address this, we construct our training set by sampling data from a literature database presented in \cite{tabor2019mapping}. Specifically, we use results for 115 paraquinone molecules as our training set and reserve 1,000 quinone molecules for testing. A crucial step in building molecular property models is featurization, which involves selecting a suitable representation of molecular structure for computational analysis. 
For this purpose, we use the open-source \texttt{PaDEL} package \cite{padel, padel-package} to compute 1,875 molecular descriptors for each molecule. These descriptors range from basic features, such as atom counts and molecular weight, to more complex graph-based properties. There are a total of 1444 one- and two-dimensional descriptors and 431 three-dimensional descriptors. Before moving to the modeling phase, we first compute the variance of all descriptors over the training set and remove those below a small threshold, leaving us with 1445 possible descriptors.
Given the high dimensionality of this problem ($d=1445$), traditional SISSO is not applicable. To manage this, we employ the mutual information (MI) screening approach in \texttt{TorchSISSO}, using the setting \texttt{initial\_screening = ["mi", 0.01]} to retain only the top 1\% of descriptors by MI value, which reduces the feature set to 11 out of the original 1,445. For comparison, we evaluate \texttt{TorchSISSO} against \texttt{VS-SISSO} \cite{VS-SISSO}, an extension of SISSO designed for high-dimensional problems that uses pre-screening. Note that \texttt{VS-SISSO} relies on the original \texttt{FORTRAN-SISSO} code for backend computations, making it a useful benchmark for our case study. No additional features were pre-screened beyond those with near zero variance, so \texttt{VS-SISSO} has access to the original 1,445 descriptors. We used the default settings for \texttt{VS-SISSO}, with two minor modifications: (i) we set the maximum number of base features \texttt{n\_max} to be 11 to match that allowed for the MI method and to ensure the computational time was reasonable and (ii) we increased the maximum number of iterations \texttt{nstep\_max} to 200 to allow increased exploration of the feature space.

\begin{figure}[tb]
    \centering
    \includegraphics[width=0.45\textwidth]{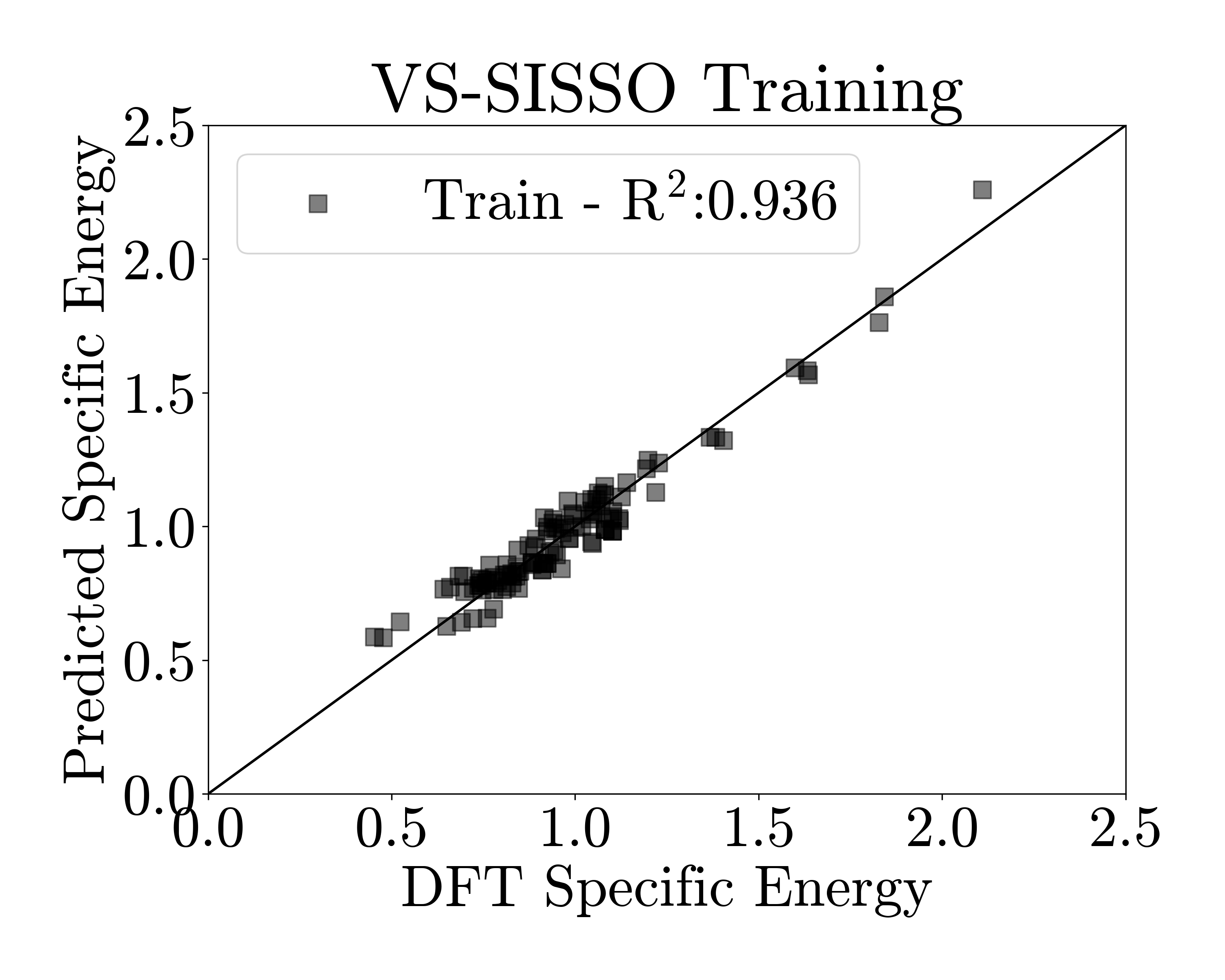}    \includegraphics[width=0.45\textwidth]{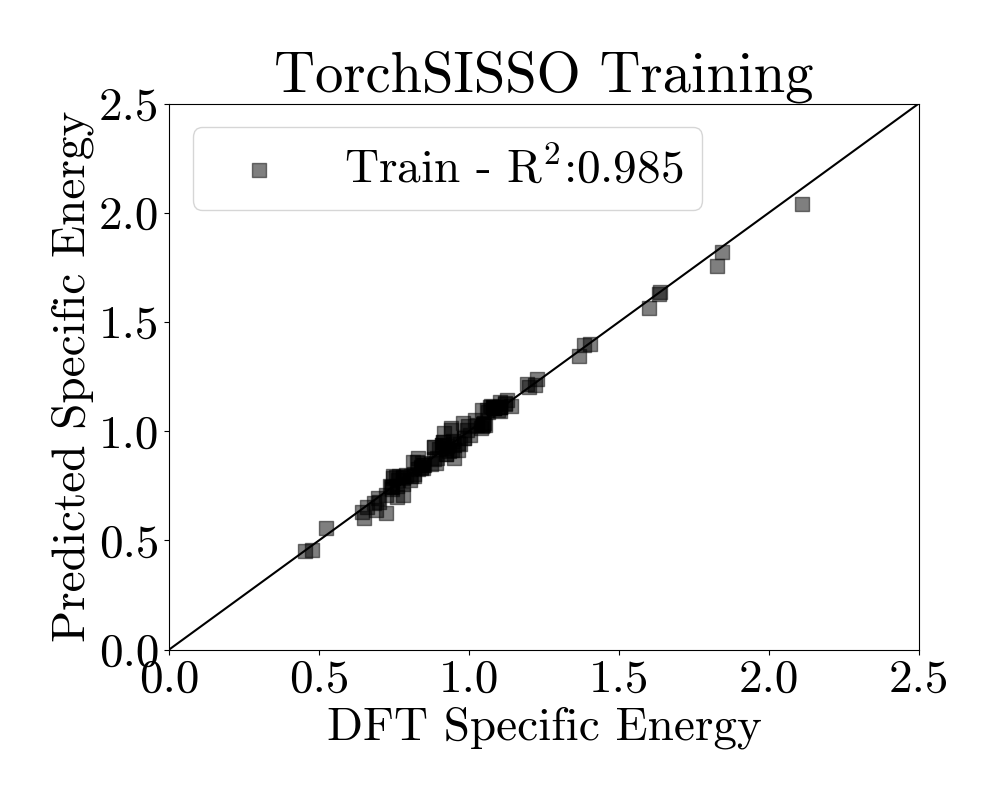}
    \includegraphics[width=0.45\textwidth]{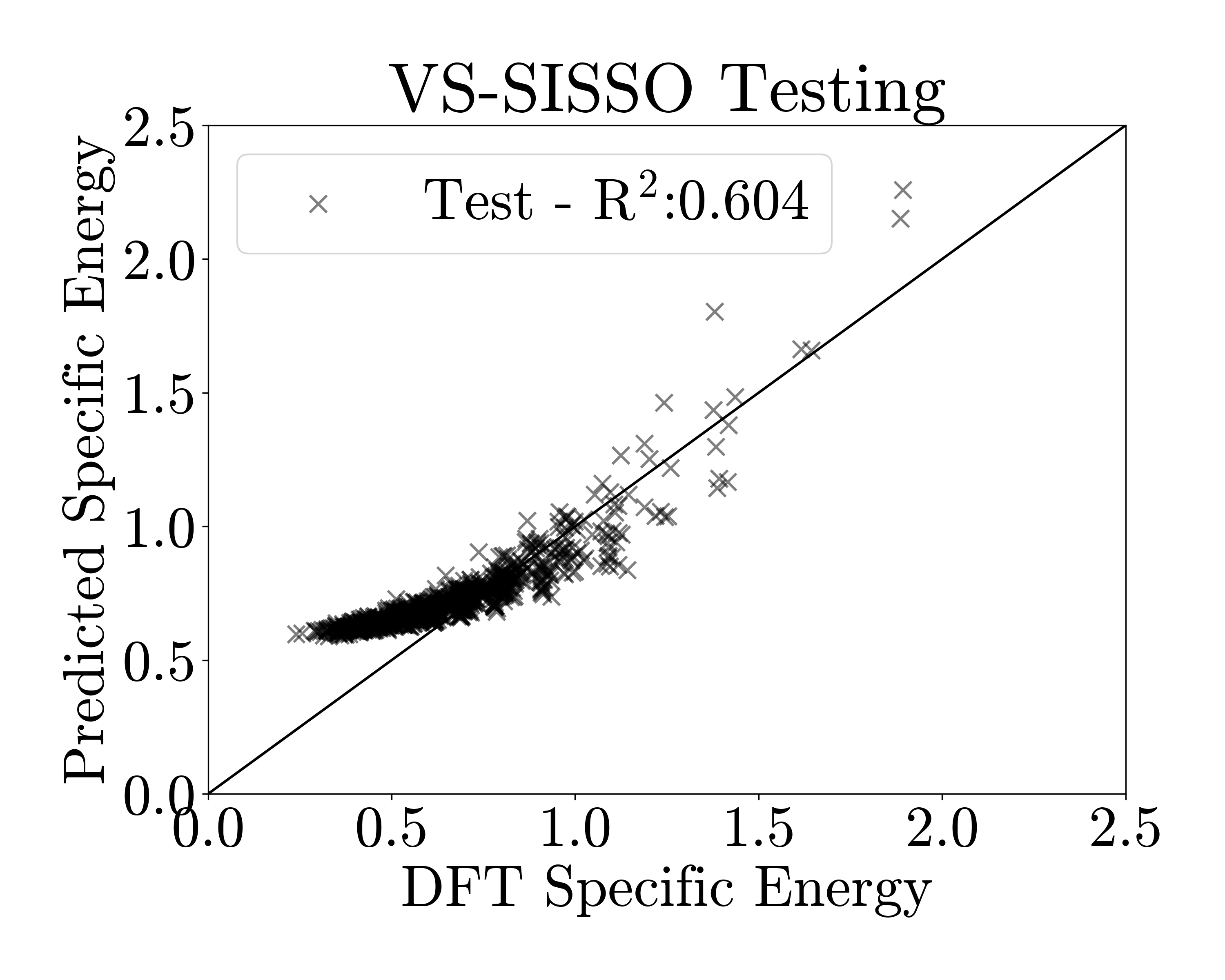}    \includegraphics[width=0.45\textwidth]{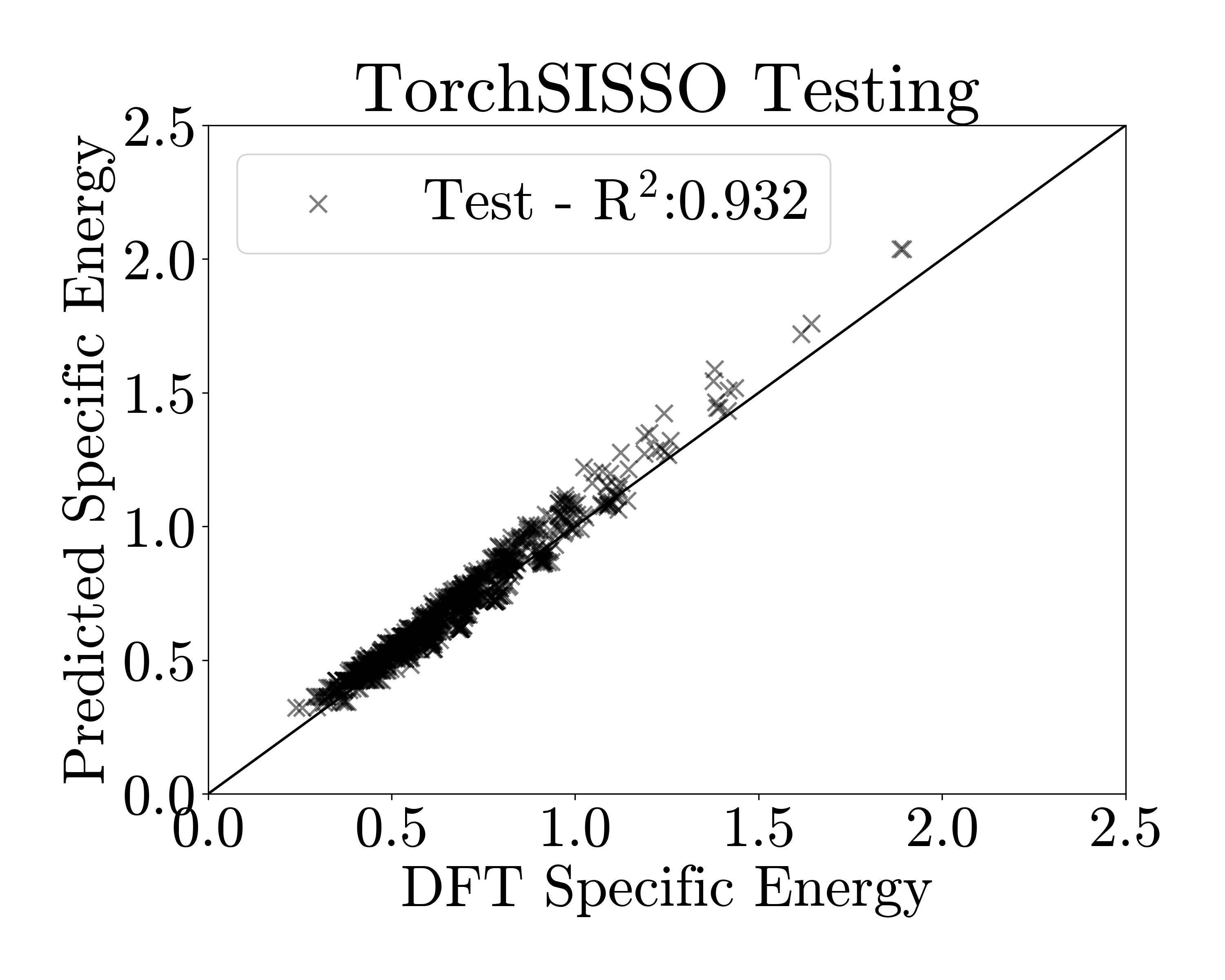}
    \caption{Results for \texttt{TorchSISSO} and \texttt{VS-SISSO} on training (top) and testing (bottom) datasets for modeling specific energy of organic compounds.}
    \label{fig:se_results}
\end{figure}

The training and testing results for both \texttt{TorchSISSO} and \texttt{VS-SISSO} are shown in Figure \ref{fig:se_results}. We see that both approaches are able to obtain good training performance, with \texttt{TorchSISSO} and \texttt{VS-SISSO} achieving $R^2$ values of 0.985 and 0.936, respectively. However, we see a bigger difference on the test data wherein \texttt{TorchSISSO} and \texttt{VS-SISSO} achieve $R^2$ values of 0.932 and 0.604, respectively. In particular, \texttt{VS-SISSO} shows a significant drop in performance for specific energy values below 0.75 where it clearly has a biased over-prediction in this range. \texttt{TorchSISSO}, on the other hand, has a much tighter parity plot throughout the full range of specific energy values, implying it has learned an equation that generalizes much better beyond than the training dataset. The CPU time required by \texttt{TorchSISSO} was just 14.3 seconds, representing an over 16-fold reduction compared to the 237.1 seconds needed for \texttt{VS-SISSO}, while also yielding a model with better predictive accuracy.

The equation found by \texttt{TorchSISSO} can be expressed as follows:
\begin{align} \label{eq:se-pred}
    \text{Specific Energy} \approx 144.14676 \left( \frac{P_{GH} + \lambda_{M4}}{M_W \times P_{GH}} \right) + 0.06388,
\end{align}
where $P_{GH}$ is the solute gas-hexadecane partition coefficient, $\lambda_{M4}$ is the largest absolute eigenvalue of the Burden modified matrix weighted by relative mass with modification parameter $n=4$, and $M_W$ is molecular weight. One interesting thing to notice right away is that \eqref{eq:se-pred} has exactly the same $M_W$ term in the denominator as \eqref{eq:specific-energy} -- we emphasize that this structure was not imposed during the training process, but was uncovered directly from the data. Although the other two features $P_{GH}$ and $\lambda_{M4}$ are not quite as intuitive, they do carry physical significance. For example, $P_{GH}$ provides a measure of how a molecule interacts with solvents, which can impact the electronic properties (such as redox potential). Despite starting with a large and complex set of potential descriptors, \texttt{TorchSISSO} was able to pinpoint a compact, interpretable equation that relies on just three fundamental molecular features, combined in a simple form, to achieve high predictive accuracy on both the training and test sets. Furthermore, the specific implementation choices clearly result in an improvement over the state-of-the-art \texttt{VS-SISSO} code for at least this real-world example. 

Note that the best (lowest error) equation found by \texttt{VS-SISSO} can be expressed as:
\begin{align}
    \text{Specific Energy} \approx 1.03252 \left( \frac{\text{ATS3i}}{\lambda_{M1}} \right) + 0.5533907820,
\end{align}
where ATS3i is the Broto-Moreau autocorrelation with lag 3 weighted by the first ionization potential and $\lambda_{M1}$ is the largest absolute eigenvalue of the Burden modified matrix weighted by relative mass with modification parameter $n=1$. Interestingly, \texttt{VS-SISSO} found a different Burden eigenvalue descriptor than \texttt{TorchSISSO}, which appears in the denominator as opposed to the numerator in the overall learned descriptor. \texttt{VS-SISSO} also does not identify the importance of molecular weight in the expression, which might be the source of the biased predictions for molecules with low specific energy values. 

\subsection{Illustration of Potential Challenges with Extrapolation}
\label{subsec:extrapolation}

In this section, we examine how the distribution of input data affects SISSO’s performance, particularly when extrapolating beyond a narrow data range. We aim to highlight challenges that arise from limited training distributions, an issue driven by data quality rather than the SISSO method itself. As an example, we use an unknown ground-truth equation based on a modified Arrhenius form:
\begin{align} \label{eq:kinetic-cons}
    k = 2.37 \sqrt{T} e^{-\frac{E_a}{RT}},
\end{align}
where $k$ is the rate constant, $T$ the temperature, $E_a$ the activation energy, and $R$ the ideal gas constant. We simulate noisy observations from \eqref{eq:kinetic-cons}, where $y = k + \varepsilon$ with $\varepsilon \sim \mathcal{N}(0,0.1)$, across various temperature ranges. For simplicity, we set \( E_a = 185 \) kJ/mol and \( R = 8.314 \times 10^{-3} \) kJ/mol-K, aiming to recover \eqref{eq:kinetic-cons} using SISSO. We assume that $E_a$ is known such that our input features include \( E_a \) and \( RT \), and the operator set \(\mathcal{O} = \{ \sqrt{(\cdot)}, (\cdot) + (\cdot), \exp(-(\cdot)), (\cdot)/(\cdot) \} \).

\begin{figure}[tb]
    \centering
    \includegraphics[width=0.49\textwidth]{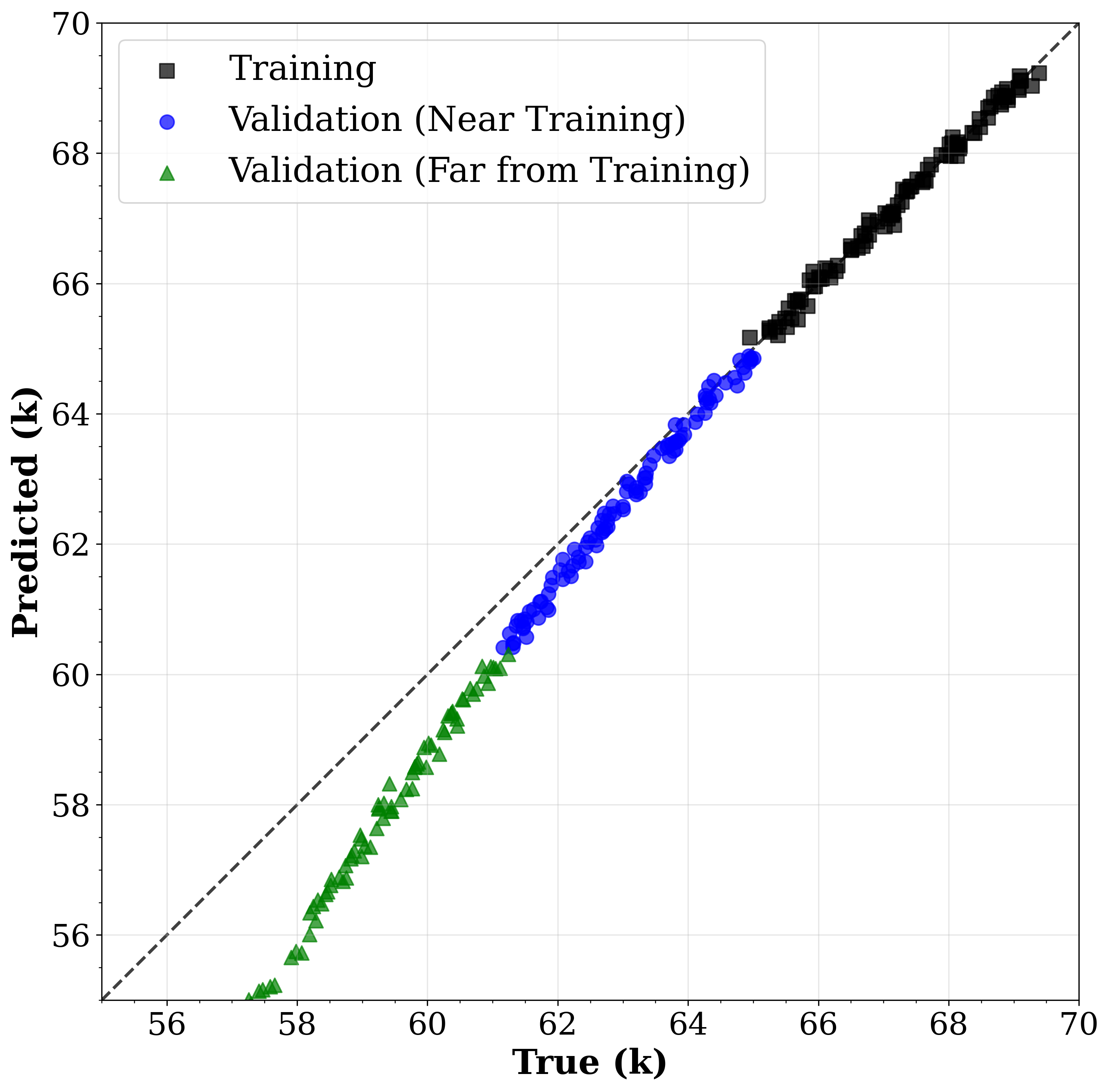}
    \includegraphics[width=0.49\textwidth]{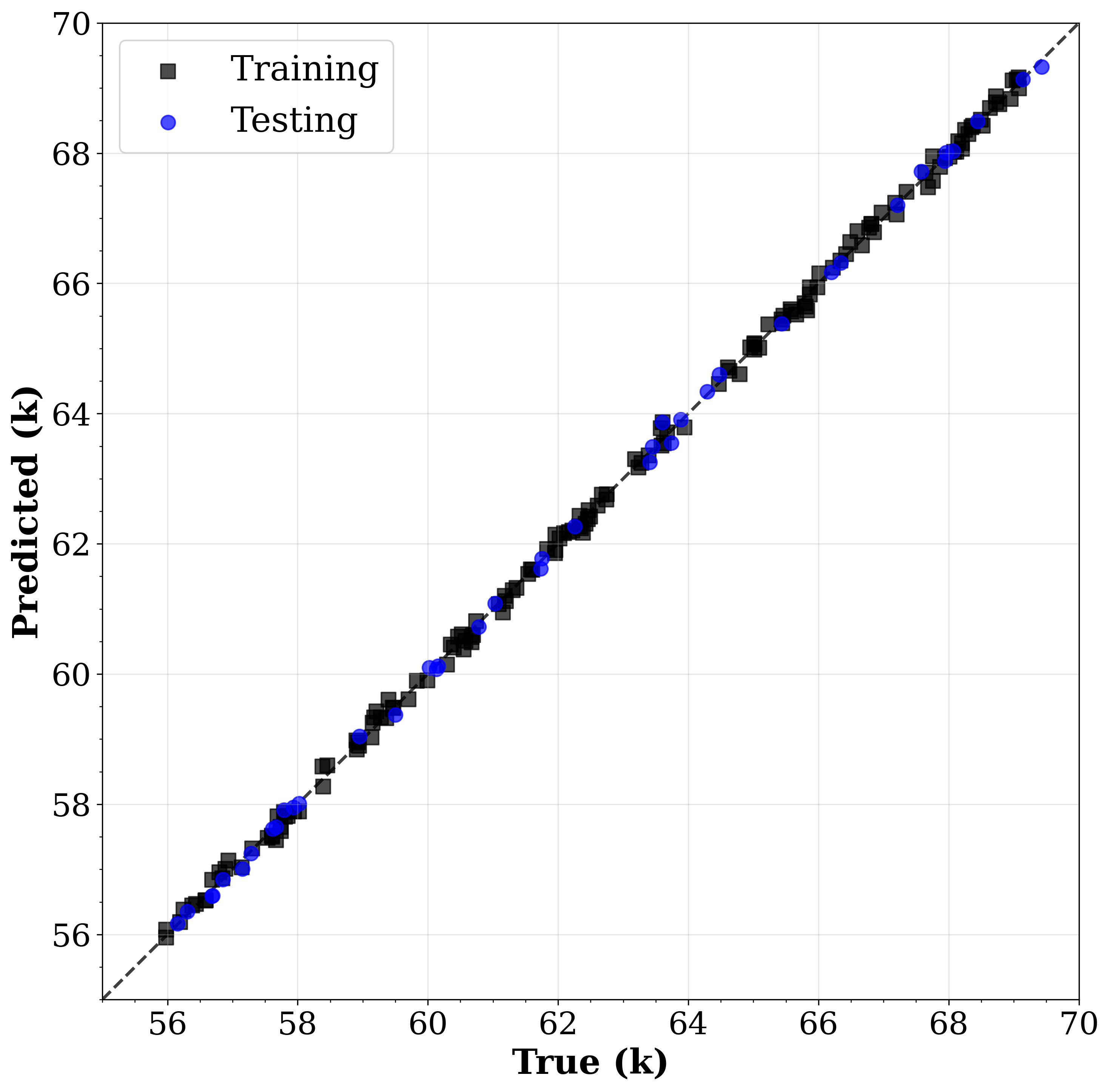}  
    \caption{(Left) Parity plot showing training and validation results for a SISSO model trained on data collected over a limited temperature range, with validation data spanning temperatures both near and far outside this range. (Right) Parity plot showing training and validation results for a SISSO model trained on data spanning the full temperature range, with an 80/20 train/validation split. This comparison illustrates the impact of training data distribution on model generalizability across a broader temperature spectrum.}
    \label{fig:limited-range}
\end{figure}

First, we train SISSO on 100 randomly generated samples from a ``training range'' of \( T \in [800,900] \) K. The resulting SISSO model after training is:
\begin{align} \label{eq:sisso-limited-kinetic}
    k \approx 1385.95 e^{-\frac{E_a}{RT}} - 1282.82,
\end{align}
which achieves a training $R^2$ of $\sim0.99$. We validate this model on two extrapolation datasets: one in a range \( T \in [710,795] \) K near that of the training data and another in a farther range of \( T \in [600,700] \) K. Figure \ref{fig:limited-range} (left) shows the parity plots, indicating that despite high accuracy within the training range, the model deviates significantly from ground truth as \( T \) moves outside this range. This is expected, as the learned model structure does not fully capture the temperature dependency of the frequency factor. The close alignment of \eqref{eq:sisso-limited-kinetic} to observed data within the training range masks its limitations, illustrating the need for rigorous testing across varied inputs, especially outside the training range.

Upon observing these deviations, one practical approach is to augment the training set with data that spans a wider temperature range. Doing so here enables SISSO to recover exactly the ground-truth structure in \eqref{eq:kinetic-cons}. Figure \ref{fig:limited-range} (right) shows the parity plot for a model trained on an 80\%/20\% training/validation split over this expanded range. The resulting SISSO model is:
\begin{align}
    k \approx 2.3723 \sqrt{T} e^{-\frac{E_a}{RT}} - 0.0752,
\end{align}
which correctly captures the frequency factor’s temperature dependence due to the improved data distribution.

In summary, while it may not always be possible to recover the exact structure of the ground-truth equation, this limitation does not prevent the effective use of learned models. Through systematic validation, we can rigorously assess model performance and identify any limitations in predictive capability. Additionally, by intentionally holding out data points near the extremes of the input space, we can further probe the model’s extrapolation capacity. Such testing methods enable us to pinpoint areas where the model may struggle. When models demonstrate consistent performance across these held-out extremes, we can be more confident in their ability to generalize beyond the training range. This approach allows us to leverage these models with greater assurance, even when the true underlying structure remains only partially known.

\subsection{Computational Time Comparison on CPU and GPU Hardware}
\label{subsec:gpu-comparison}

To further demonstrate the advantages of \texttt{TorchSISSO} on hardware accelerators, specifically GPUs, we compare runtime performance between \texttt{TorchSISSO} and \texttt{FORTRAN-SISSO} across several computing configurations. While the default version of \texttt{FORTRAN-SISSO} is limited to single-core processors, a recent update introduces multi-core support. However, due to the complex installation requirements for multi-core usage, we could not implement this on our cluster, so this configuration is excluded from the comparisons. A key advantage of \texttt{TorchSISSO} is its versatility: it can be readily installed and executed on single-core and multi-core CPUs, as well as on GPUs, and we examine all these configurations in this section. We used an NVIDIA A100 Tensor Core GPU with 40 GB of RAM on our computing cluster, capping CPU memory to 40 GB to ensure a consistent maximum RAM across all tests.

\begin{figure}[tb]
    \centering
    \includegraphics[width=0.8\textwidth]{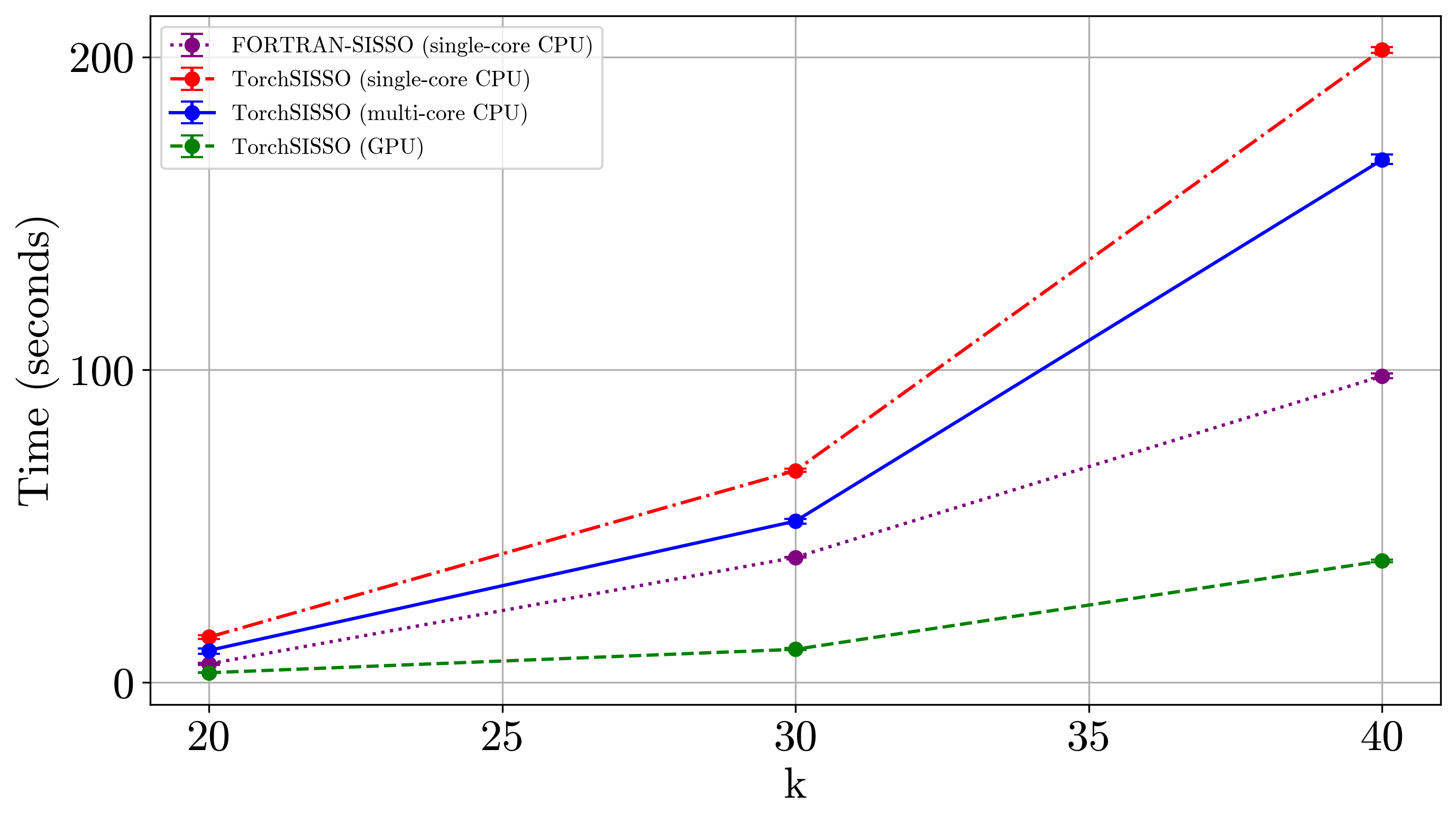}        \caption{Computational time versus parameter $k$ for running \texttt{TorchSISSO} and \texttt{FORTRAN-SISSO} on different hardware. Note that $k$ controls how many models must be fit according to \eqref{eq:sis-method}--\eqref{eq:subspace-expansion}.}
    \label{fig:time-results}
\end{figure}

To assess runtime performance on more complex regression tasks, we constructed three four-term models: (i) $x_1^4 + x_2^3 + x_3^2 + x_4$; (ii) $x_1^5 + x_2^4 + x_3^2 + x_4$; and (iii) $x_1^3 + x_2^2 + x_3 + \sin(x_4)$. For each model, 100 data points were generated, with each input component sampled uniformly over $[1,5]$. Execution times for \texttt{FORTRAN-SISSO} (single CPU core) and \texttt{TorchSISSO} (single- and multi-core CPU, and GPU) as a function of the parameter $k$ are illustrated in Figure \ref{fig:time-results}. Error bars reflect minimum and maximum times across the three test cases. As expected, training times increase with $k$, which determines the number of features selected per term for regression and, hence, the number of models fit. Across all cases, GPU runtime is the shortest, with \texttt{FORTRAN-SISSO} as the second fastest, likely due to advantages inherent in optimized Fortran compilers for scientific computing. It is worth noting that \texttt{TorchSISSO} was slightly faster than \texttt{FORTRAN-SISSO} on most of the previous benchmarks (Section \ref{subsec:synthetic}--\ref{subsec:scientific}) since the equations of interest were not as complicated as that considered in this section and we did not need to cap the CPU memory to fairly compare against the GPU. Moreover, we used default, non-optimized settings for \texttt{TorchSISSO} in all of our runs, so further optimization has the potential to yield even greater speed improvements. Despite the lack of any systematic optimization, the GPU results show a substantial (2-4x) reduction in time compared to the optimized Fortran implementation.

Finally, we assess performance for ensemble model training. Using the first four-term model, we trained 30 different models by partitioning the data into 30 subsets of 100 samples each (fixing $k=20$ to its default value). In the torch framework, all training tasks can be dispatched simultaneously to the GPU, enabling substantial time savings compared to sequential CPU and Fortran implementations, as shown in Table \ref{tab:gpu-ensemble}.

\begin{table}[tb]
        \centering
        \caption{Computational times (seconds) for training an ensemble of 30 models each with 100 datapoints for a four term ground-truth expression using \texttt{FORTRAN-SISSO} and \texttt{TorchSISSO} under different hardware configurations.}
        \begin{tabular}{cccc} \hline
        \texttt{FORTRAN-SISSO} &\texttt{TorchSISSO} (single-CPU) &\texttt{TorchSISSO} (multi-CPU) &\texttt{TorchSISSO} (GPU) \\ \hline 
        182.43 &321.87 &287.34 &51.32 \\
        \hline
        \end{tabular}
\label{tab:gpu-ensemble}
\end{table}

\section{Conclusions} \label{sec:conclusion}

In this work, we introduced \texttt{TorchSISSO}, a native Python implementation of the Sure Independence Screening and Sparsifying Operator (SISSO) method, designed to overcome the limitations of the original FORTRAN-based implementation. By leveraging the PyTorch framework, \texttt{TorchSISSO} provides enhanced flexibility, allowing users to easily modify the feature expansion process and integrate modern computational resources such as GPUs for significant speed-ups. This adaptability removes barriers to install
ation and usage, particularly in cloud-based or high-performance computing environments, making the SISSO method accessible to a broader scientific community.

Our results demonstrate that \texttt{TorchSISSO} performs comparably or better than the original SISSO implementation across a range of tasks, including synthetic test equations, scientific benchmarks, and real-world applications such as molecular property prediction. Notably, \texttt{TorchSISSO} shows improved accuracy in discovering true symbolic expressions in cases where the original \texttt{FORTRAN-SISSO} implementation falters. Additionally, the reduction in computational time, achieved through parallel processing and optional GPU acceleration, makes \texttt{TorchSISSO} a highly scalable tool for symbolic regression tasks on larger datasets and more complex feature spaces.

In summary, \texttt{TorchSISSO} addresses the key limitations of the original SISSO method, offering a faster, more accessible, and more adaptable solution for symbolic regression across a wide range of scientific fields. We believe this tool will facilitate the discovery of interpretable models in materials science, physics, and beyond, while also empowering researchers to further customize the method to fit specific domain needs. Future work will focus on extending the functionality of \texttt{TorchSISSO}, including multi-objective optimization, advanced regularization techniques, and automated hyperparameter tuning to further enhance its applicability.

\section*{Acknowledgements}

The authors gratefully acknowledge financial support from the National Science Foundation under Grant No. 2237616. 

\appendix
\numberwithin{equation}{section}
\numberwithin{figure}{section}
\numberwithin{table}{section}

\section{Appendix: Optimal Expressions for Benchmark Problems}

In this appendix, we summarize the exact expressions found by \texttt{TorchSISSO} and \texttt{FORTRAN-SISSO} on the synthetic and scientific benchmark problems. Note that \texttt{TorchSISSO} found the correct model structure in all cases and only shows some relatively minor discrepancy in the constants due to the noise present in the training data. 

\subsection{Synthetic Equations}

The expressions found by \texttt{TorchSISSO} and \texttt{FORTRAN-SISSO} on the synthetic case study equations, corresponding to the results in Table \ref{tab:synthetic}, are summarized below in Table \ref{tab:synthetic-equations-found}. 

\begin{table}[tb]
        \centering
        \caption{Expressions found with \texttt{TorchSISSO} and \texttt{FORTRAN-SISSO} for the synthetic case studies reported in Table \ref{tab:synthetic}. The bold numbers represent cases where \texttt{FORTRAN-SISSO} failed to identify the correct equation.}
        \begin{tabular}{ccc} \hline
        \# &\texttt{TorchSISSO} &\texttt{FORTRAN-SISSO} \\ \hline 
        1 & $10.068\frac{x_1}{x_2(x_3 + x_4)}$ & $10.068\frac{x_1}{x_2(x_3 + x_4)}$ \\[2mm]
        2 & $2.11\sin(x_2) + 3.007\sqrt{x_1}$ &$2.11\sin(x_2) + 3.007\sqrt{x_1}$ \\[2mm]
        \textbf{3} &$3.09\frac{\exp(x_1)}{x_2 + \exp(x_3)}$ & $2.065 \frac{x_3\exp(x_1)}{x_2}$ \\[2mm]
        4 &$0.99x_1^3 + 0.99x_2^2 + 2.99x_3$ &$0.99x_1^3 + 0.99x_2^2 + 2.99x_3$ \\[2mm]
        \textbf{5} & $0.98 \frac{x_2 + \exp(x_2)}{x_1^2 - x_2^2}$ & $0.53\frac{x_1 \exp(x_1)}{x_1 - x_2}$ \\[2mm]
        6 &$\sqrt{x_1^2 + x_2^2} + 0.018$ & $\sqrt{x_1^2 + x_2^2} + 0.018$ \\[2mm]
        7 & $0.98\sin(x_1 x_3) + 1.49\exp(-x_1 x_2) - 0.0176$ & $0.98\sin(x_1 x_3) + 1.49\exp(-x_1 x_2) - 0.0176$ \\[2mm]
        8 & $4.99 x_1 x_3^2 + x_1^3 + 3x_1 x_2^2$ & $4.99 x_1 x_3^2 + x_1^3 + 3x_1 x_2^2$ \\[2mm]
        \textbf{9} & $x_1 x_2 x_3 ( \ln(x_4) - \ln(x_5) )$ & $1.42 x_3^2 \left( \frac{x_4 - x_5}{x_4} \right) + 0.31$ \\[2mm]
        10 & $\exp(-\frac{x_1}{x_2 x_3})$ & $\exp(-\frac{x_1}{x_2 x_3})$ \\ \hline
        \end{tabular}
\label{tab:synthetic-equations-found}
\end{table}

\subsection{Scientific Equations}

The expressions found by \texttt{TorchSISSO} and \texttt{FORTRAN-SISSO} on the scientific benchmarks, corresponding to the results in Table \ref{tab:equations_phys}, are summarized below in Table \ref{tab:scientific-equations-found}. 

\begin{table}[tb]
        \centering
        \caption{Expressions found with \texttt{TorchSISSO} and \texttt{FORTRAN-SISSO} for the scientific benchmarks reported in Table \ref{tab:equations_phys}. The bold names represent cases where \texttt{FORTRAN-SISSO} failed to identify the correct equation.}
        \begin{tabular}{ccc} \hline
        Name &\texttt{TorchSISSO} &\texttt{FORTRAN-SISSO} \\\hline
        \textbf{Distance} & $d^2 = (x_0-x_1)^2+(x_2-x_3)^2$ & $d^2 = 1.843(x_1-x_2)^4 + 1.45$  \\[2mm]
        \textbf{Particle Displacement} & $F=q(E + Bv\sin(\theta))$ & $F = 0.742 q \theta (E + B_ v) + 0.13$ \\[2mm]
        \textbf{Relativistic Mass} & $m^2 = \frac{m_0^2 }{1 - \frac{v_1^2}{c^2}}$ & $m^2 = \frac{m_0^2 c - v}{c} - 0.967$ \\[2mm]
        \textbf{Oscillation Amplitude} & $x = \frac{qe} { m(\omega_1^2-\omega_2^2)}$ & $x = -0.978\left( \frac{e}{m \omega_1 \omega_2} \right) - 0.24$ \\\hline
        \end{tabular}
\label{tab:scientific-equations-found}
\end{table}

\bibliographystyle{unsrt}  
\bibliography{ref}

\end{document}